\documentclass[10pt,twocolumn,letterpaper]{article}

\pdfoutput=1

\usepackage{cvpr}
\usepackage{times}
\usepackage{epsfig}
\usepackage{graphicx}
\usepackage{amsmath}
\usepackage{amssymb}
\usepackage{array,booktabs}
\usepackage[linesnumbered,ruled,vlined]{algorithm2e}
\usepackage{multirow}

\allowdisplaybreaks[4]

\usepackage[pagebackref=true,breaklinks=true,letterpaper=true,colorlinks,bookmarks=false]{hyperref}

\cvprfinalcopy 

\def\cvprPaperID{2661} 

\begin{document}

\title{Simultaneous Super-Resolution and Cross-Modality Synthesis of 3D Medical Images using Weakly-Supervised Joint Convolutional Sparse Coding}

\author{Yawen Huang$^1$,
	Ling Shao$^2$,
	Alejandro F. Frangi$^1$\\
	$^1$Department of Electronic and Electrical Engineering, The University of Sheffield, UK\\
	$^2$School of Computing Sciences, University of East Anglia, UK\\
	{\tt\small \{yhuang36, a.frangi\}@sheffield.ac.uk, ling.shao@uea.ac.uk}
}

\maketitle

\begin{abstract}
	Magnetic Resonance Imaging (MRI) offers high-resolution \emph{in vivo} imaging and rich functional and anatomical multimodality tissue contrast. In practice, however, there are challenges associated with considerations of scanning costs, patient comfort, and scanning time that constrain how much data can be acquired in clinical or research studies. In this paper, we explore the possibility of generating high-resolution and multimodal images from low-resolution single-modality imagery. We propose the weakly-supervised joint convolutional sparse coding to simultaneously solve the problems of super-resolution (SR) and cross-modality image synthesis. The learning process requires only a few registered multimodal image pairs as the training set. Additionally, the quality of the joint dictionary learning can be improved using a larger set of unpaired images\footnote{Unpaired data/images: acquisitions are from different subjects without registration. Paired data/images: acquisitions of the same subject obtained from different modalities are registered.}. To combine unpaired data from different image resolutions/modalities, a hetero-domain image alignment term is proposed. Local image neighborhoods are naturally preserved by operating on the whole image domain (as opposed to image patches) and using joint convolutional sparse coding. The paired images are enhanced in the joint learning process with unpaired data and an additional maximum mean discrepancy term, which minimizes the dissimilarity between their feature distributions. Experiments show that the proposed method outperforms state-of-the-art techniques on both SR reconstruction and simultaneous SR and cross-modality synthesis.
\end{abstract}

\section{Introduction}
With the rapid progress in Magnetic Resonance Imaging (MRI), there are a multitude of mechanisms to generate tissue contrast that are associated with various anatomical or functional features. However, the acquisition of a complete multimodal set of high-resolution images faces constraints associated with scanning costs, scanner availability, scanning time, and patient comfort. In addition, long-term longitudinal studies such as ADNI~\cite{mueller2005alzheimer} imply that changes exist in the scanner or acquisition protocol over time. In these situations, it is not uncommon to have images of the same subject but obtained from different sources, or to be confronted with missing or corrupted data from earlier time points. In addition, high-resolution (HR) 3D medical imaging usually requires long breath-hold and repetition times, which lead to long-term scanning times that are challenging or unfeasible in clinical routine. Acquiring low-resolution (LR) images and/or skipping some imaging modalities altogether from the acquisition are then not uncommon. In all such scenarios, it is highly desirable to be able to generate HR data from the desired target modality from the given LR modality data.

The relevant literature in this area can be divided into either super-resolution (SR) reconstruction from single/multiple image modalities or cross-modality (image) synthesis (CMS). On the one hand, SR is typically concerned with achieving improved visual quality or overcoming the resolution limits of the acquired image data. Such a problem is generally under-determined and ill-posed, hence, the solution is not unique. To mitigate this fact, the solution space needs to be constrained by incorporating strong priors. Prior information comes in the form of smoothness assumptions as in, for example, interpolation-based SR~\cite{keys1981cubic,shao2007order}. State-of-the-art methods mostly adopt either external data or internal data to guide the learning algorithms~\cite{rousseau2010non,tang2017pairwise}. On the other hand, due to variations in optimal image representations across modalities, the learned image model from one modality data may not be the optimal model for a different modality. How to reveal the relationship between different representations of the underlying image information is a major research issue to be explored. In order to synthesize one modality from another, recent methods in CMS proposed utilizing non-parametric methods like nearest neighbor (NN) search~\cite{freeman2000learning}, nonlinear regression forests~\cite{jog2013magnetic}, coupled dictionary learning~\cite{roy2013magnetic}, and convolutional neural network (CNN)~\cite{gatys2016image}, to name a few. Although these algorithms achieve remarkable results, most of them suffer from the fundamental limitations associated with supervised learning and/or patch-based synthesis. Supervised approaches require a large number of training image pairs, which is impractical in many medical imaging applications. Patch-based synthesis suffers from inconsistencies introduced during the fusion process that takes place in areas where patches overlap.

In this paper, we propose a weakly-supervised convolutional sparse coding method with an application to neuroimaging that utilizes a small set of registered multimodal image pairs and solves the SR and CMS problems simultaneously. Rather than factorizing each patch into a linear combination of patches drawn from a dictionary built under sparsity constraints (sparse coding), or requiring a training set with fully registered multimodal image pairs, or requiring the same sparse code to be used for both modalities involved, we generate a unified learning model that automatically learns a joint representation for heterogeneous data (\eg, different resolutions, modalities and relative poses). This representation is learned in a common feature space that preserves the local consistency of the images. Specifically, we utilize the co-occurrence of texture features across both domains. A manifold ranking method picks features of the target domain from the most similar subjects in the source domain. Once the correspondence between images in different domains is established, we directly work on a whole image representation that intrinsically respects local neighborhoods. Furthermore, a mapping function is learned that links the representations between the two modalities involved. We call the proposed method \underline{WE}akly-sup\underline{E}rvised joi\underline{N}t convolut\underline{I}onal spars\underline{E} coding (WEENIE), and perform extensive experiments to verify its performance.

The main contributions of this paper are as follows: 1) This is the first attempt to jointly solve the SR and CMS problems in 3D medical imaging using weakly-supervised joint convolutional sparse coding; 2) To exploit unpaired images from different domains during the learning phase, a hetero-domain image alignment term is proposed, which allows identifying correspondences across source and target domains and is invariant to pose transformations; 3) To map LR and HR cross-modality image pairs, joint learning based on convolutional sparse coding is proposed that includes a maximum mean discrepancy term; 4) Finally, extensive experimental results show that the proposed model yields better performance than state-of-the-art methods in both reconstruction error and visual quality assessment measures.
\vspace*{-2mm}
\section{Related Work}
\vspace*{-2mm}
With the goal to transfer the modality information from the source domain to the target domain, recent developments in CMS, such as texture synthesis~\cite{efros2001image,gatys2016image,hertzmann2001image}, face photo-sketch synthesis~\cite{gao2012face,wang2009face}, and multi-modal retrieval~\cite{monay2007modeling,smeulders2000content}, have shown promising results. In this paper, we focus on the problems of image super-resolution and cross-modality synthesis, so only review related methods on these two aspects.

\textbf{Image Super-Resolution}: The purpose of image SR is to reconstruct an HR image from its LR counterpart. According to the image priors, image SR methods can be grouped into two main categories: interpolation-based, external or internal data driven learning methods. Interpolation-based SR works, including the classic bilinear~\cite{li2001new}, bicubic~\cite{keys1981cubic}, and some follow-up methods~\cite{shao2007order,zhang2006edge}, interpolate much denser HR grids by the weighted average of the local neighbors. Most modern image SR methods have shifted from interpolation to learning based. These methods focus on learning a compact dictionary or manifold space to relate LR/HR image pairs, and presume that the lost high-frequency (HF) details of LR images can be predicted by learning from either external datasets or internal self-similarity. The external data driven SR approaches~\cite{chang2004super,freeman2002example,yang2010image} exploit a mapping relationship between LR and HR image pairs from a specified external dataset. In the pioneer work of Freeman \etal~\cite{freeman2002example}, the NN of an LR patch is found, with the corresponding HR patch, and used for estimating HF details in a Markov network. Chang \etal~\cite{chang2004super} projected multiple NNs of the local geometry from the LR feature space onto the HR feature space to estimate the HR embedding. Furthermore, sparse coding-based methods~\cite{rueda2013single,yang2010image} were explored to generate a pair of dictionaries for LR and HR patch pairs to address the image SR problem. Wang \etal~\cite{wang2012semi} and Huang \etal~\cite{huang2013coupled} further suggested modeling the relationship between LR and HR patches in the feature space to relax the strong constraint. Recently, an efficient CNN based approach was proposed in~\cite{dong2016image}, which directly learned an end-to-end mapping between LR and HR images to perform complex nonlinear regression tasks. For internal dataset driven SR methods, this can be built using the similarity searching~\cite{rousseau2010non} and/or scale-space pyramid of the given image itself~\cite{huang2015single}.

\textbf{Cross-Modality Synthesis}: In parallel, various CMS methods have been proposed for synthesizing unavailable modality data from available source images, especially in the medical imaging community~\cite{roy2013magnetic,van2015cross,vemulapalli2015unsupervised}. One of the well-established modality transformation approaches is the example-based learning method generated by Freeman \etal~\cite{freeman2000learning}. Given a patch of a test image, several NNs with similar properties are picked from the source image space to reconstruct the target one using Markov random fields. Roy \etal~\cite{roy2013magnetic} used sparse coding for desirable MR contrast synthesis assuming that cross-modality patch pairs have same representations and can be directly used for training dictionaries to estimate the contrast of the target modality. Similar work was also used in~\cite{iglesias2013synthesizing}. In~\cite{bahrami2015hierarchical}, a canonical correlation analysis-based approach was proposed to yield a feature space that can get underlying common structures of co-registered data for better correlation of dictionary pairs. More recently, a location-sensitive deep network~\cite{van2015cross} has been put forward to explicitly utilize the voxel image coordinates by incorporating image intensities and spatial information into a deep network for synthesizing purposes. Gatys \etal~\cite{gatys2016image} introduced a CNN algorithm of artistic style, that new images can be generated by performing a pre-image search in high-level image content to match generic feature representations of example images. In addition to the aforementioned methods, most CMS algorithms rely on the strictly registered pairs to train models. As argued in~\cite{vemulapalli2015unsupervised}, it would be preferable to use an unsupervised approach to deal with input data instead of ensuring data to be coupled invariably. 
\begin{figure*}
	\setlength{\abovecaptionskip}{0.cm}
	\setlength{\belowcaptionskip}{-0.5cm}
	\centering
	\includegraphics[width=0.8\linewidth]{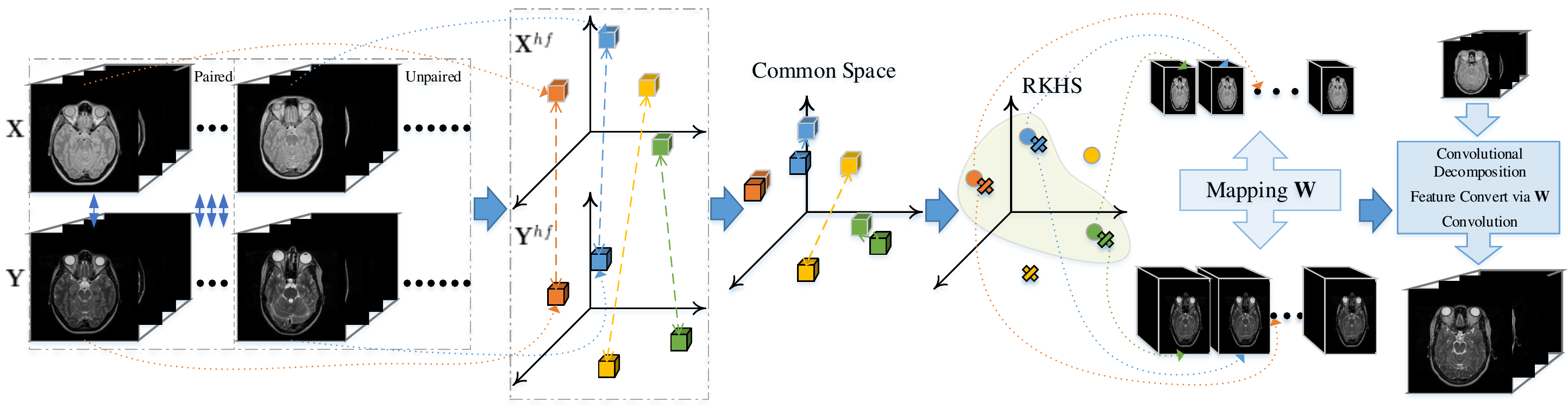}
	\vspace*{-1mm}
	\caption{Flowchart of the proposed method (WEENIE) for simultaneous SR and cross-modality synthesis.}
	\label{fig1}
	\vspace*{-5mm}
\end{figure*}
\section{Weakly-Supervised Joint Convolutional Sparse Coding}
\subsection{Preliminaries}
Convolutional Sparse Coding (CSC) was introduced in the context of modeling receptive fields preciously, and later generalized to image processing, in which the representation of an entire image is computed by the sum of a set convolutions with dictionary filters. The goal of CSC is to remedy the shortcoming of conventional patch-based sparse coding methods by removing shift variations for consistent approximation of local neighbors on whole images. Concretely, given the vectorized image $\mathbf{x}$, the problem of generating a set of vectorized filters for sparse feature maps is solved by minimizing the objective function that combines the squared reconstruction error and the $l_1$-norm penalty on the representations:
\begin{equation}
\begin{aligned}
\label{eq１}
\arg \min_{\mathbf{f},\mathbf{z}} \frac{1}{2}\left \| \mathbf{x}-\sum_{k=1}^{K}\mathbf{f}_{k}\ast \mathbf{z}_{k} \right \|_{2}^{2}+\lambda \sum_{k=1}^{K}\left \| \mathbf{z}_{k} \right \|_{1}\\
s.t. \, \left \| \mathbf{f}_{k} \right \|_{2}^{2}\leq 1 \; \forall k=\left \{ 1,...,K \right \},\\
\end{aligned}
\end{equation}
where $\mathbf{x}$ is an $m \times n$ image in vector form, $\mathbf{f}_{k}$ refers to the $k$-th $d \times d$ filter in vector form, $\mathbf{z}_{k}$ is the sparse feature map corresponding to $\mathbf{f}_{k}$ with size $\left ( m+d-1 \right )\times \left ( n+d-1 \right )$ to approximate $\mathbf{x}$, $\lambda$ controls the $l_1$ penalty, and $\ast$ denotes the 2D convolution operator. $\mathbf{f}=\left [ \mathbf{f}_{1}^{T},...,\mathbf{f}_{K}^{T} \right ]^{T}$ and $\mathbf{z}=\left [ \mathbf{z}_{1}^{T},...,\mathbf{z}_{K}^{T} \right ]^{T}$ are $K$ filters and feature maps stacked as the single column vector, respectively. Here, the inequality constraint on each column of vectorized $\mathbf{f}_{k}$ prevents the filter from absorbing all the energy of the system. 

Similar to the original sparse coding problem, Zeiler \etal~\cite{zeiler2010deconvolutional} proposed to solve the CSC in Eq. (\ref{eq１}) through alternatively optimizing one variable while fixing the other one in the spatial domain. Advances in recent fast convolutional sparse coding (FCSC)~\cite{bristow2013fast} have shown that feature learning can be efficiently and explicitly solved by incorporating CSC within an alternating direction method of multipliers (ADMMs) framework in the Fourier domain.

\subsection{Problem Formulation}
The simultaneous SR and cross-modality synthesis problem can be formulated as: given a three-dimensional LR image $\mathbf{X}$ of modality $\mathcal{M}_{1}$, the task is to infer from $\mathbf{X}$ a target 3D image $\mathbf{Y}$ that is as similar as possible to the HR ground truth of desirable modality $\mathcal{M}_{2}$. Suppose that we are given a group of LR images of modality $\mathcal{M}_{1}$, \ie, $\mathcal{X}=\left [ \mathbf{X}_{1}, ...,\mathbf{X}_{P}\right ] \in \mathbb{R}^{m\times n\times t\times P}$, and a set of HR images of modality $\mathcal{M}_{2}$, \ie, $\mathcal{Y}=\left [ \mathbf{Y}_{2}, ...,\mathbf{Y}_{Q}\right ] \in \mathbb{R}^{m\times n\times t\times Q}$. $P$ and $Q$ are the numbers of samples in the training sets, and $m$, $n$ denote the dimensions of axial view of each image, while $t$ is the size of the image along the z-axis. Moreover, in both training sets, subjects of source modality $\mathcal{M}_{1}$ are mostly different from target modality $\mathcal{M}_{2}$, that is, we are working with a small number of paired data while most of them are unpaired. Therefore, the difficulties of this problem vary with hetero-domain images, \eg, resolutions and modalities, and how well the two domains fit. To bridge image appearances across heterogeneous representations, we propose a method for automatically establishing a one-to-one correlation between data in $\mathcal{X}$ and $\mathcal{Y}$ firstly, then employ the aligned data to jointly learn a pair of filters, while assuming that there exists a mapping function $\mathcal{F}\left ( \cdot  \right )$ for associating and predicting cross-modality data in the projected common feature space. Particularly, we want to synthesize MRI of human brains in this paper. An overview of our proposed work is depicted in Fig. \ref{fig1}.

\textbf{Notation}: For simplicity, we denote matrices and 3D images as upper-case bold (\eg, image $\mathbf{X}$), vectors and vectorized 2D images as lower-case bold (\eg, filter $\mathbf{f}$), and scalars as lower-case (\eg, the number of filter $k$). Image with modality $\mathcal{M}_{1}$ called source modality belongs to the source domain, and with modality $\mathcal{M}_{2}$ called target modality belongs to the target domain. 

\subsection{Hetero-Domain Image Alignment}
The design of an alignment $\mathcal{A}\left ( \cdot  \right )$ from $\mathcal{X}$ to $\mathcal{Y}$ requires a combination of extracting common components from LR/HR images and some measures of correlation between both modalities. In SR literature, common components are usually accomplished by extracting high-frequency (HF) edges and texture features from LR/HR images, respectively~\cite{chang2004super,yang2010image}. In this paper, we adopt first- and second-order derivatives involving
horizontal and vertical gradients as the features for LR images by $\mathbf{X}_{p}^{hf}=\mathcal{G} \ast \mathbf{X}_{p}$. $\mathcal{G}=\begin{bmatrix} \mathbf{G}_{1}^{1},\, \mathbf{G}_{1}^{2}\\ \mathbf{G}_{2}^{1},\, \mathbf{G}_{2}^{2} \end{bmatrix}$, and each gradient $\mathbf{G}$ has the same length of z-axis as input image while $\mathbf{g}_{1}^{1}=[-1,0,1]$, $\mathbf{g}_{1}^{2}={\mathbf{g}_{1}^{1}}^{T}$, and $\mathbf{g}_{2}^{1}=[-2,-1,0,1,2]$, $\mathbf{g}_{2}^{2}={\mathbf{g}_{2}^{1}}^{T}$. For HR images, HF features are obtained through directly subtracting mean value, \ie, $\mathbf{Y}_{p}^{hf}=\mathbf{Y}_{p}-mean(\mathbf{Y}_{p})$. To define the hetero-domain image alignment term $\mathcal{A}\left ( \cdot  \right )$, we assume that the intrinsic structures of brain MRI of a subject across image modalities are also similar in the HF space since images of different modalities are more likely to be described differently by features. When HF features of both domains are obtained, it is possible to build a way for cross-modality data alignment (in particular, a unilateral cross-modality matching can be thought as a special case in~\cite{huang2016color}). To this end, we define a subject-specific transformation matrix $\mathbb{A}$ as
\begin{equation}
\begin{aligned}
\label{eq2}
\mathbb{A}=\begin{bmatrix}
K(\mathbf{X}_{1}^{hf},\mathbf{Y}_{1}^{hf})& \cdots & K(\mathbf{X}_{1}^{hf},\mathbf{Y}_{Q}^{hf})\\ 
\vdots & \ddots & \vdots \\ 
K(\mathbf{X}_{P}^{hf},\mathbf{Y}_{1}^{hf}) & \cdots & K(\mathbf{X}_{P}^{hf},\mathbf{Y}_{Q}^{hf})
\end{bmatrix},\\
\end{aligned}
\end{equation}
where $K(\mathbf{X}_{p}^{hf},\mathbf{Y}_{q}^{hf})$ is used for measuring the distances between each pair of HF data in $\mathcal{X}$ and $\mathcal{Y}$ computed by the Gaussian kernel as
\begin{equation}
\begin{aligned}
\label{eq3}
K(\mathbf{X}_{p}^{hf},\mathbf{Y}_{q}^{hf})=\frac{1}{(\sqrt{2\pi}\sigma)^{3}}e^{-\frac{\left | \mathbf{X}_{p}^{hf} - \mathbf{Y}_{q}^{hf} \right |^{2}}{2\sigma^{2}}},\\
\end{aligned}
\end{equation}
where $\sigma$ determines the width of Gaussian kernel. In order to establish a one-to-one correspondence across different domains, for each element of $\mathcal{X}$, the most relevant image with maximum $K$ from $\mathcal{Y}$ is preserved while discarding the rest of the elements:
\begin{equation}
\begin{aligned}
\label{eq4}
\mathbb{A}=\begin{bmatrix}
\max \left ( K\left(1,:\right) \right ) & & \\ 
& \ddots & \\ 
& & \max \left ( K\left(P,:\right) \right )
\end{bmatrix},\\
\end{aligned}
\end{equation}
where $\max \left ( K\left(p,:\right) \right )$ denotes the maximum element of the $p$-th row of $\mathbb{A}$. We further set $\max \left ( K\left(p,:\right) \right )$ to 1, and all the blank elements to 0. Therefore, $\mathbb{A}$ is a binary matrix. Since $\mathbb{A}$ is calculated in a subject-specific manner, each subject of $\mathcal{X}$ can only be connected to one target of the most similar brain structures. Hence, images under a hetero-domain can be treated as being the registered pairs, \ie, $\mathcal{P}_{i}=\left \{\mathbf{X}_{i}, \mathbf{Y}_{i}\right \}_{i=1}^{P}$, by constructing virtual correspondence: $\mathcal{A}(\mathcal{X},\mathcal{Y})=\left \| \mathbf{X}^{hf}- \mathbb{A}\mathbf{Y}^{hf}\right \|_{2}^{2}$.

\subsection{Objective Function}
For image modality transformation, coupled sparse coding~\cite{jia2013image,yang2010image} has important advantages, such as reliability of correspondence dictionary pair learning and less memory cost. However, the arbitrarily aligned bases related to the small part of images may lead to shifted versions of the same structures or inconsistent representations based on the overlapped patches. CSC~\cite{zeiler2010deconvolutional} was then proposed to generate a global decomposition framework based on the whole image for solving the above problem. In spired by CSC and the benefits of coupled sparsity~\cite{jia2013image}, we introduce a joint convolutional sparse coding method in a weakly-supervised setting for hetero-domain images. The small number of originally registered pairs are used to carry the intrinsic relationship between $\mathcal{X}$ and $\mathcal{Y}$ while the majority of unpaired data are introduced to exploit and enhance the diversity of the original learning system. 

Assume that the aforementioned alignment approach leads to a perfect correspondence across $\mathcal{X}$ and $\mathcal{Y}$, such that each aligned pair of images possesses approximately identical (or the same for co-registered data) information. Moreover, to facilitate image mappings in a joint manner, we require sparse feature maps of each pair of corresponding source and target images to be associated. That is, suppose that there exists a mapping function $\mathcal{F}\left ( \cdot  \right )$, where the feature maps of LR $\mathcal{M}_{1}$ modality images can be converted to their HR $\mathcal{M}_{2}$ versions. Given $\mathcal{X}$ and $\mathcal{Y}$, we propose to learn a pair of filters with corresponding feature maps and a mapping function together with the aligned term by
\begin{equation}
\begin{aligned}
\label{eq5}
\arg \min_{\mathbf{F}^{x},\mathbf{F}^{y},\mathbf{Z}^{x},\mathbf{Z}^{y}, \mathbf{W}} \frac{1}{2} \left \| \mathbf{X}-\sum_{k=1}^{K}\mathbf{F}_{k}^{x}\ast \mathbf{Z}_{k}^{x} \right \|_{F}^{2}\\
+\frac{1}{2} \left \| \mathbf{Y}-\sum_{k=1}^{K}\mathbf{F}_{k}^{y}\ast \mathbf{Z}_{k}^{y} \right \|_{F}^{2}+\beta\sum_{k=1}^{K}\left \| \mathbf{Z}_{k}^{y}-\mathbf{W}_{k}\mathbf{Z}_{k}^{x} \right \|_{F}^{2}\\
+\lambda\left ( \sum_{k=1}^{K}\left \| \mathbf{Z}_{k}^{x} \right \|_{1} +\sum_{k=1}^{K}\left \| \mathbf{Z}_{k}^{y} \right \|_{1} \right )+\gamma \sum_{k=1}^{K}\left \| \mathbf{W}_{k} \right \|_{F}^{2}\\
+\left \| \mathbf{X}^{hf}- \mathbb{A}\mathbf{Y}^{hf}\right \|_{2}^{2} \,\, s.t. \, \left \| \mathbf{f}_{k}^{x} \right \|_{2}^{2}\leq 1, \left \| \mathbf{f}_{k}^{y} \right \|_{2}^{2}\leq 1 \,\forall k,\\
\end{aligned}
\end{equation}
where $\mathbf{Z}_{k}^{x}$ and $\mathbf{Z}_{k}^{y}$ are the $k$-th sparse feature maps that estimate the aligned data terms $\mathbf{X}$ and $\mathbf{Y}$ when convolved with the $k$-th filters $\mathbf{F}_{k}^{x}$ and $\mathbf{F}_{k}^{y}$ of a fixed spatial support, $\forall k=\left \{ 1,...,K \right \}$. Concretely, $\mathbf{X}$ denotes the aligned image from $\mathcal{P}$ with LR and $\mathcal{M}_{1}$ modality; $\mathbf{Y}$ denotes the aligned image from $\mathcal{P}$ containing HR and $\mathcal{M}_{2}$ modality. A convolution operation is represented as $\ast$ operator, and $\left \| \cdot  \right \|_{F}$ denotes a Frobenius norm chosen to induce the convolutional least squares approximate solution. $\mathbf{F}^{x}$ and $\mathbf{F}^{y}$ are adopted to list all $K$ filters, while $\mathbf{Z}^{x}$ and $\mathbf{Z}^{y}$ represent corresponding $K$ feature maps for source and target domains, respectively. $\mathcal{A}\left ( \mathcal{X},\mathcal{Y} \right )$ is combined to enforce the correspondence for unpaired auxiliary subjects. The mapping function $\mathcal{F}\left ( \mathbf{Z}_{k}^{x}, \mathbf{W}_{k}\right )=\mathbf{W}_{k} \mathbf{Z}_{k}^{x}$ is modeled as a linear projection $\mathbf{W}_{k}$ of $\mathbf{Z}_{k}^{x}$ and $\mathbf{Z}_{k}^{y}$ by solving a set of the least squares problem (\ie, $\min_\mathbf{W} \sum_{k=1}^{K}\left \| \mathbf{Z}_{k}^{y}-\mathbf{W}_{k}\mathbf{Z}_{k}^{x} \right \|_{F}^{2}$). Parameters $\lambda$, $\beta$ and $\gamma$ balance sparsity, feature representation and association mapping.

It is worth noting that $\mathcal{P}_{i}=\left \{\mathbf{X}_{i}, \mathbf{Y}_{i}\right \}$ may not be perfect since HF feature alignment in Eq. (\ref{eq4}) is not good enough for very heterogeneous domain adaptation by matching the first- and second-order derivatives of $\mathcal{X}$ and means of $\mathcal{Y}$, which leads to suboptimal filter pairs and inaccurate results. To overcome such a problem, we need additional constraints to ensure the correctness of registered image pairs produced by the alignment. Generally, when feature difference is substantially large, there always exists some subjects of the source domain that are not particularly related to target ones even in the HF subspace. Thus, a registered subject pairs' divergence assessment procedure should be cooperated with the aforementioned joint learning model to handle this difficult setting. Recent works~\cite{chen2014recognizing,long2013transfer,zheng2016hetero} have performed instance/domain adaptation via measuring data distribution divergence using the maximum mean discrepancy (MMD) criterion. We follow such an idea and employ the empirical MMD as the nonparametric distribution measure to handle the hetero-domain image pair mismatch problem in the reproducing kernel Hilbert space (RKHS). This is done by minimizing the difference between distributions of aligned subjects while keeping dissimilar 'registered' pairs (\ie, discrepant distributions) apart in the sparse feature map space:
\begin{equation}
\begin{aligned}
\label{eq6}
\frac{1}{P}\sum_{i=1}^{P}\sum_{k=1}^{K}\left \| \mathbf{W}_{k}(i)\mathbf{Z}_{k}^{x}(i)-\mathbf{Z}_{k}^{y}(i) \right \|_{\mathcal{H}}^{2}\\
=\sum_{k=1}^{K}(\mathbf{W}_{k}\mathbf{Z}_{k}^{x})^{T}M_{i}\mathbf{Z}_{k}^{y}=Tr(\sum_{k=1}^{K}\mathbf{Z}_{k}^{y}\mathbf{M}(\mathbf{W}_{k}\mathbf{Z}_{k}^{x})^{T}),\\
\end{aligned}
\end{equation}
where $\mathcal{H}$ indicates RKHS space, $\mathbf{Z}_{k}^{x}(i)$ and $\mathbf{Z}_{k}^{y}(i)$ are the paired sparse feature maps for $\mathcal{P}_{i}=\{\mathbf{X}_{i},\mathbf{Y}_{i}\}$ with $i=1,...P$, $M_{i}$ is the $i$-th element of $\mathbf{M}$ while $\mathbf{M}$ denotes the MMD matrix and can be computed as follows
\begin{equation}
\begin{aligned}
\label{eq7}
M_{i}=\left\{\begin{matrix}
&\frac{1}{P}, \; &\mathbf{Z}_{k}^{x}(i),\mathbf{Z}_{k}^{y}(i) \in \mathcal{P}_{i}\\ 
&-\frac{1}{P^{2}}, \; &\mathrm{otherwise}.
\end{matrix}\right.,\\
\end{aligned}
\end{equation}

By regularizing Eq. (\ref{eq5}) with Eq. (\ref{eq6}), filter pairs $\mathbf{F}_{k}^{x}$ and $\mathbf{F}_{k}^{y}$ are refined and the distributions of real aligned subject pairs are drawn close under the new feature maps. Putting the above together, we obtain the objective function:
\begin{equation}
\begin{aligned}
\label{eq:obj}
\arg \min_{\mathbf{F}^{x},\mathbf{F}^{y},\mathbf{Z}^{x},\mathbf{Z}^{y}, \mathbf{W}} \frac{1}{2} \left \| \mathbf{X}-\sum_{k=1}^{K}\mathbf{F}_{k}^{x}\ast \mathbf{Z}_{k}^{x} \right \|_{F}^{2}+\gamma \sum_{k=1}^{K}\left \| \mathbf{W}_{k} \right \|_{F}^{2}\\
+\frac{1}{2} \left \| \mathbf{Y}-\sum_{k=1}^{K}\mathbf{F}_{k}^{y}\ast \mathbf{Z}_{k}^{y} \right \|_{F}^{2}+\beta\sum_{k=1}^{K}\left \| \mathbf{Z}_{k}^{y}-\mathbf{W}_{k}\mathbf{Z}_{k}^{x} \right \|_{F}^{2}\\
+\lambda\left ( \sum_{k=1}^{K}\left \| \mathbf{Z}_{k}^{x} \right \|_{1} +\sum_{k=1}^{K}\left \| \mathbf{Z}_{k}^{y} \right \|_{1} \right )+Tr(\sum_{k=1}^{K}\mathbf{Z}_{k}^{y}\mathbf{M}(\mathbf{W}_{k}\mathbf{Z}_{k}^{x})^{T})\\
+\left \| \mathbf{X}^{hf}- \mathbb{A}\mathbf{Y}^{hf}\right \|_{2}^{2} \; s.t. \, \left \| \mathbf{f}_{k}^{x} \right \|_{2}^{2}\leq 1, \left \| \mathbf{f}_{k}^{y} \right \|_{2}^{2}\leq 1 \,\forall k.\\
\end{aligned}
\end{equation}

\subsection{Optimization}
We propose a three-step optimization strategy for efficiently tackling the objective function in Eq. (\ref{eq:obj}) (termed (WEENIE), summarized in Algorithm \ref{alg1}) considering that such multi-variables and unified framework cannot be jointly convex to $\mathbf{F}$, $\mathbf{Z}$, and $\mathbf{W}$. Instead, it is convex with respect to each of them while fixing the remaining variables. 

\subsubsection{Computing Convolutional Sparse Coding}
Optimization involving only sparse feature maps $\mathbf{Z}^{x}$ and $\mathbf{Z}^{y}$ is solved by initialization of filters $\mathbf{F}^{x}$, $\mathbf{F}^{y}$ and mapping function $\mathbf{W}$ ($\mathbf{W}$ is initialized as an identity matrix). Besides the original CSC formulation, we have additional terms associated with data alignment and divergence reducing in the common feature space. Eq. (\ref{eq:obj}) is firstly converted to two regularized sub-CSC problems. Unfortunately, each of the problems constrained with an $l_{1}$ penalty term cannot be directly solved, which is not rotation invariant. Recent approaches~\cite{bristow2013fast,heide2015fast} have been proposed to work around this problem on the theoretical derivation by introducing two auxiliary variables $\mathbf{U}$ and $\mathbf{S}$ to enforce the constraint inherent in the splitting. To facilitate component-wise multiplications, we exploit the convolution subproblem~\cite{bristow2013fast} in the Fourier domain\footnote{Fast Fourier transform (FFT) is utilized to solve the relevant linear system and demonstrated substantially better asymptotic performance than processed in the spatial domain.} derived within the ADMMs framework:
\begin{equation}
\begin{aligned}
\notag
\min_{\mathbf{Z}^{x}}\frac{1}{2} \left \| \hat{\mathbf{X}}-\sum_{k=1}^{K}\hat{\mathbf{F}}_{k}^{x}\odot  \hat{\mathbf{Z}}_{k}^{x} \right \|_{F}^{2}+\left \| \mathbf{X}^{hf}-\mathbb{A}\mathbf{Y}^{hf}\right \|_{2}^{2}\\
+Tr(\sum_{k=1}^{K}\hat{\mathbf{Z}}_{k}^{y}\mathbf{M}(\mathbf{W}_{k}\hat{\mathbf{Z}}_{k}^{x})^{T})+\beta\sum_{k=1}^{K}\left \| \hat{\mathbf{Z}}_{k}^{y}-\mathbf{W}_{k}\hat{\mathbf{Z}}_{k}^{x} \right \|_{F}^{2}\\
+\lambda\sum_{k=1}^{K}\left \| \mathbf{U}_{k}^{x} \right \|_{1} \,\, s.t.\, \left \| \mathbf{S}_{k}^{x} \right \|_{2}^{2}\leq 1,\mathbf{S}_{k}^{x}=\boldsymbol{\Phi}^{T}\hat{\mathbf{F}}_{k}^{x},\mathbf{U}_{k}^{x}=\mathbf{Z}_{k}^{x} \;\forall k,\\ 
\end{aligned}
\end{equation}
\begin{equation}
\begin{aligned}
\label{eq9}
\min_{\mathbf{Z}^{y}}\frac{1}{2} \left \| \hat{\mathbf{Y}}-\sum_{k=1}^{K}\hat{\mathbf{F}}_{k}^{y}\odot  \hat{\mathbf{Z}}_{k}^{y} \right \|_{F}^{2}+\left \| \mathbf{X}^{hf}-\mathbb{A}\mathbf{Y}^{hf}\right \|_{2}^{2}\\
+Tr(\sum_{k=1}^{K}\hat{\mathbf{Z}}_{k}^{y}\mathbf{M}(\mathbf{W}_{k}\hat{\mathbf{Z}}_{k}^{x})^{T})+\beta\sum_{k=1}^{K}\left \| \hat{\mathbf{Z}}_{k}^{x}-\mathbf{W}_{k}\hat{\mathbf{Z}}_{k}^{y} \right \|_{F}^{2}\\
+\lambda\sum_{k=1}^{K}\left \| \mathbf{U}_{k}^{y} \right \|_{1} \,\, s.t.\, \left \| \mathbf{S}_{k}^{y} \right \|_{2}^{2}\leq 1,\mathbf{S}_{k}^{y}=\boldsymbol{\Phi}^{T}\hat{\mathbf{F}}_{k}^{y},\mathbf{U}_{k}^{y}=\mathbf{Z}_{k}^{y} \;\forall k,
\end{aligned}
\end{equation}
where $\hat{}$ applied to any symbol indicates the discrete Fourier transform (DFT), for example $\hat{\mathbf{X}} \leftarrow f(\mathbf{X})$, and $f(\cdot)$ denotes the Fourier transform operator. $\odot$ represents the Hadamard product (\ie, component-wise product), $\boldsymbol{\Phi}^{T}$ is the inverse DFT matrix, and $s$ projects a filter onto a small spatial support. By utilizing slack variables $\mathbf{U}_{k}^{x}$, $\mathbf{U}_{k}^{y}$ and $\mathbf{S}_{k}^{x}$, $\mathbf{S}_{k}^{y}$, the loss function can be treated as the sum of multiple subproblems and with the addition of equality constraints.

\subsubsection{Training Filters}
Similar to theoretical CSC methods, we alternatively optimize the convolutional least squares term for the basis function pairs $\mathbf{F}^{x}$ and $\mathbf{F}^{y}$ followed by an $l_{1}$-regularized least squares term for the corresponding sparse feature maps $\mathbf{Z}^{x}$ and $\mathbf{Z}^{y}$. Like the subproblem of solving feature maps, filter pairs can be learned in a similar fashion. With $\hat{\mathbf{Z}}_{k}^{x}$, $\hat{\mathbf{Z}}_{k}^{y}$ and $\mathbf{W}_{k}$ fixed, we can update the corresponding filter pairs $\hat{\mathbf{F}}_{k}^{x}$, and $\hat{\mathbf{F}}_{k}^{y}$ as
\begin{equation}
\begin{aligned}
\label{eq10}
\min_{\mathbf{F}^{x},\mathbf{F}^{y}} \frac{1}{2} \left \| \hat{\mathbf{X}}-\sum_{k=1}^{K}\hat{\mathbf{F}}_{k}^{x}\odot \hat{\mathbf{Z}}_{k}^{x} \right \|_{F}^{2}+\frac{1}{2} \left \| \hat{\mathbf{Y}}-\sum_{k=1}^{K}\hat{\mathbf{F}}_{k}^{y}\odot \hat{\mathbf{Z}}_{k}^{y} \right \|_{F}^{2}\\
s.t. \, \left \| \mathbf{f}_{k}^{x} \right \|_{2}^{2}\leq 1, \left \| \mathbf{f}_{k}^{y} \right \|_{2}^{2}\leq 1 \; \forall k,\\
\end{aligned}
\end{equation}
The optimization with respect to Eq. (\ref{eq10}) can be solved by a one-by-one update strategy~\cite{wang2012semi} through an augmented Lagrangian method~\cite{bristow2013fast}.

\subsubsection{Learning Mapping Function}
Finally, $\mathbf{W}_{k}$ can be learned by fixing $\mathbf{F}_{k}^{x}$, $\mathbf{F}_{k}^{y}$, and $\mathbf{Z}_{k}^{x}$, $\mathbf{Z}_{k}^{y}$:
\begin{equation}
\begin{aligned}
\label{eq11}
\min_{\mathbf{W}}\sum_{k=1}^{K}\left \| \mathbf{Z}_{k}^{y}-\mathbf{W}_{k}\mathbf{Z}_{k}^{x} \right \|_{F}^{2}+\left (\frac{\gamma}{\beta}  \right )\sum_{k=1}^{K}\left \| \mathbf{W}_{k} \right \|_{F}^{2}\\
+Tr(\sum_{k=1}^{K}\mathbf{Z}_{k}^{y}\mathbf{M}(\mathbf{W}_{k}\mathbf{Z}_{k}^{x})^{T}),\\
\end{aligned}
\end{equation}
where Eq. (\ref{eq11}) is a ridge regression problem with a regularization term. We simplify the regularization term $\mathcal{R}(tr)=Tr(\sum_{k=1}^{K}\mathbf{Z}_{k}^{y}\mathbf{M}(\mathbf{W}_{k}\mathbf{Z}_{k}^{x})^{T})$ and analytically derive the solution as $\mathbf{W}=(\mathbf{Z}_{k}^{y}{\mathbf{Z}_{k}^{x}}^{T}-\mathcal{R}(tr))(\mathbf{Z}_{k}^{x}{\mathbf{Z}_{k}^{x}}^{T}+\frac{\gamma}{\beta} \mathbf{I})^{-1}$, where $\mathbf{I}$ is an identity matrix.

\begin{algorithm}[h]
	\caption{WEENIE Algorithm}
	\label{alg1}
	\SetAlgoLined
	\KwIn{Training data $\mathbf{X}$ and $\mathbf{Y}$, parameters $\lambda$, $\gamma$, $\sigma$.}
	Initialize $\mathbf{F}_{0}^{x}$, $\mathbf{F}_{0}^{y}$, $\mathbf{Z}_{0}^{x}$, $\mathbf{Z}_{0}^{y}$, $\mathbf{U}_{0}^{x}$, $\mathbf{U}_{0}^{y}$, $\mathbf{S}_{0}^{x}$, $\mathbf{S}_{0}^{y}$, $\mathbf{W}_{0}$.\\
	Perform FFT $\mathbf{Z}_{0}^{x}\rightarrow \hat{\mathbf{Z}}_{0}^{x}$, $\mathbf{Z}_{0}^{y}\rightarrow \hat{\mathbf{Z}}_{0}^{y}$, $\mathbf{F}_{0}^{x}\rightarrow \hat{\mathbf{F}}_{0}^{x}$, $\mathbf{F}_{0}^{y}\rightarrow \hat{\mathbf{F}}_{0}^{y}$, $\mathbf{U}_{0}^{x}\rightarrow \hat{\mathbf{U}}_{0}^{x}$, $\mathbf{U}_{0}^{y}\rightarrow \hat{\mathbf{U}}_{0}^{y}$, $\mathbf{S}_{0}^{x}\rightarrow \hat{\mathbf{S}}_{0}^{x}$, $\mathbf{S}_{0}^{y}\rightarrow \hat{\mathbf{S}}_{0}^{y}$.\\
	Let $\hat{\mathbf{Z}}_{0}^{y} \leftarrow \mathbf{W}\hat{\mathbf{Z}}_{0}^{x}$.\\
	\While{not converged}{
		Fix other variables, update $\hat{\mathbf{Z}}_{k+1}^{x}$, $\hat{\mathbf{Z}}_{k+1}^{y}$ and $\hat{\mathbf{U}}_{k+1}^{x}$, $\hat{\mathbf{U}}_{k+1}^{y}$ by (\ref{eq9}).\\
		Fix other variables, update $\hat{\mathbf{F}}_{k+1}^{x}$, $\hat{\mathbf{F}}_{k+1}^{y}$ and $\hat{\mathbf{S}}_{k+1}^{x}$, $\hat{\mathbf{S}}_{k+1}^{y}$ by (\ref{eq10}) with $\hat{\mathbf{Z}}_{k+1}^{x}$, $\hat{\mathbf{Z}}_{k+1}^{y}$, $\hat{\mathbf{U}}_{k+1}^{x}$, $\hat{\mathbf{U}}_{k+1}^{y}$ and $\mathbf{W}_{k}$.\\
		Fix other variables, update $\mathbf{W}_{k}$ by (\ref{eq11}) with $\hat{\mathbf{Z}}_{k+1}^{x}$, $\hat{\mathbf{Z}}_{k+1}^{y}$, $\hat{\mathbf{U}}_{k+1}^{x}$, $\hat{\mathbf{U}}_{k+1}^{y}$, $\hat{\mathbf{F}}_{k+1}^{x}$, $\hat{\mathbf{F}}_{k+1}^{y}$, and $\hat{\mathbf{S}}_{k+1}^{x}$, $\hat{\mathbf{S}}_{k+1}^{y}$.\\
		Inverse FFT $\hat{\mathbf{F}}_{k+1}^{x}\rightarrow \mathbf{F}_{k+1}^{x}$, $\hat{\mathbf{F}}_{k+1}^{y}\rightarrow \mathbf{F}_{k+1}^{y}$.\\
	}
	\KwOut{$\mathbf{F}^{x}$, $\mathbf{F}^{y}$, $\mathbf{W}$.}
\end{algorithm}

\subsection{Synthesis}
Once the training stage is completed, generating a set of filter pairs $\mathbf{F}^{x}$, $\mathbf{F}^{y}$ and the mapping $\mathbf{W}$, for a given test image $\mathbf{X}^{t}$ in domain $\mathcal{X}$, we can synthesize its desirable HR version of style $\mathcal{Y}$. This is done by computing the sparse feature maps $\mathbf{Z}^{t}$ of $\mathbf{X}^{t}$ with respect to a set of filters $\mathbf{F}^{x}$, and associating $\mathbf{Z}^{t}$ to the expected feature maps $\hat{\mathbf{Z}}^{t}$ via $\mathbf{W}$, \ie,  $\hat{\mathbf{Z}}^{t} \approx \mathbf{W}\mathbf{Z}^{t}$. Therefore, the desirable HR $\mathcal{M}_{2}$ modality image is then obtained by the sum of $K$ converted sparse feature maps $\hat{\mathbf{Z}}_{k}^{t}$ convolved with desired filters $\mathbf{F}_{k}^{y}$ (termed (SRCMS) summarized in Algorithm \ref{alg2}):
\begin{equation}
\begin{aligned}
\label{eq12}
\mathbf{Y}^{t}=\sum_{k=1}^{K}\mathbf{F}_{k}^{y}\mathbf{W}_{k}{\mathbf{Z}}_{k}^{t}=\sum_{k=1}^{K}\mathbf{F}_{k}^{y}\hat{\mathbf{Z}}_{k}^{t}.\\
\end{aligned}
\end{equation}
\begin{algorithm}[h]
	\caption{SRCMS}
	\label{alg2} 
	\SetAlgoLined
	\KwIn{Test image $\mathbf{X}^{t}$, filter pairs $\mathbf{F}^{x}$ and $\mathbf{F}^{y}$, mapping $\mathbf{W}$.}
	Initialize $\mathbf{Z}_{0}^{t}$.\\
	Let $\hat{\mathbf{Z}}_{0}^{t} \leftarrow \mathbf{W}\mathbf{Z}_{0}^{t}$, $\mathbf{Y}_{0}^{t} \leftarrow \mathbf{F}^{y}\mathbf{W}\mathbf{Z}_{0}^{t}$.\\
	\While{not converged}{
		Update $\mathbf{Z}_{k+1}^{t}$ and $\hat{\mathbf{Z}}_{k+1}^{t}$ by (\ref{eq9}) with $\mathbf{Y}_{k}^{t}$, and $\mathbf{W}$.\\
		Update $\mathbf{Y}_{k+1}^{t} \leftarrow \mathbf{W}\hat{\mathbf{Z}}_{k+1}^{t}$.\\
	}
	Synthesize $\mathbf{Y}^{t}$ by (\ref{eq12}).\\
	\KwOut{Synthesized image $\mathbf{Y}^{t}$.}
\end{algorithm}

\section{Experimental Results}
\begin{figure}[t]
	\setlength{\abovecaptionskip}{0.cm}
	\setlength{\belowcaptionskip}{-0.5cm}
	\centering
	\includegraphics[width=1\linewidth]{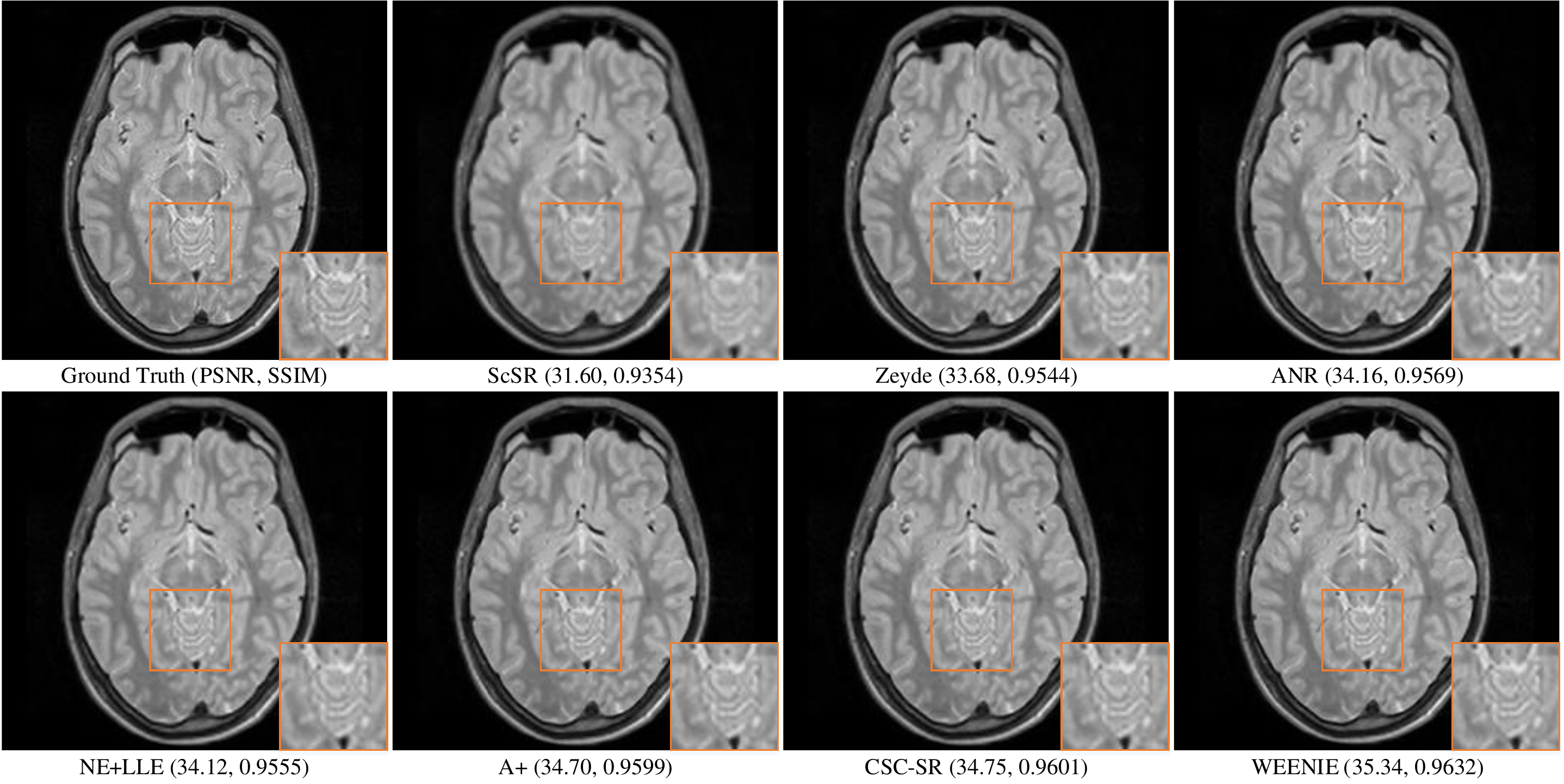}
	\vspace*{-5mm}
	\caption{Example SR results and corresponding PSNRs, SSIMs (zoom in for details).}
	\label{fig2}
	\vspace*{-4mm}
\end{figure}
We conduct the experiments using two datasets, \ie, IXI\footnote{http://brain-development.org/ixi-dataset/} and NAMIC brain mutlimodality\footnote{http://hdl.handle.net/1926/1687} datasets. Following~\cite{gu2015convolutional,wang2012semi,yang2010image}, LR counterparts are directly down-sampled from their HR ground truths with rate $1/2$ by bicubic interpolation, boundaries are padded (with eight pixels) to avoid the boundary effect of Fourier domain implementation. The regularization parameters $\sigma$, $\lambda$, $\beta$, and $\gamma$ are empirically set to be 1, 0.05, 0.1, 0.15, respectively. Optimization variables $\mathbf{F}$, $\mathbf{S}$, $\mathbf{Z}$, and $\mathbf{U}$ are randomly initialized with Gaussian noise considering~\cite{bristow2013fast}. Generally, a larger number of filters leads to better results. To balance between computation complexity and result quality, we learn 800 filters following~\cite{gu2015convolutional}.
In our experiments, we perform a more challenging division by applying half of the dataset (processed to be weakly co-registered data) for training while the remaining for testing. To the best of our knowledge, there is no previous work specially designed for SR and cross-modality synthesis simultaneously by learning from the weakly-supervised data. Thus, we extend the range of existing works as the baselines for fair comparison, which can be divided into two categories as follows: (1) brain MRI SR; (2) SR and cross-modality synthesis (one-by-one strategy in comparison models). For the evaluation criteria, we adopt the widely used PSNR and SSIM~\cite{wang2004image} indices to objectively assess the quality of the synthesized images. 

\textbf{Experimental Data}: The IXI dataset consists of 578 $256 \times 256 \times n$ MR healthy subjects collected at three hospitals with different mechanisms (\ie, Philips 3T system, Philips 1.5T system, and GE 3T system). Here, we utilize 180 Proton Density-weighted (PD-w) MRI subjects for image SR, while applying both PD-w and registered T2-weighted (T2-w) MRI scans of all subjects for major SRCMS. Further, we conduct SRCMS experiments on the processed NAMIC dataset, which consists of 20 $128 \times 128 \times m$ subjects in both T1-weighted (T1-w) and T2-w modalities. As mentioned, we leave half of the dataset out for cross-validation. We randomly select 30 registered subject pairs for IXI, and 3 registered subject pairs for NAMIC, respectively, from the half of the corresponding dataset for training purposes, and process the reminding training data to be unpaired. Particularly, all the existing methods with respect to cross-modality synthesis in brain imaging request a pre-processing, \ie, skull stripping and/or bias corrections, as done in~\cite{vemulapalli2015unsupervised,roy2013magnetic}. We follow such processes and further validate whether pre-processing (especially skull stripping) is always helpful for brain image synthesis.

\subsection{Brain MRI Super-Resolution}
\setlength{\textfloatsep}{9pt}
\begin{figure}[t]
	\setlength{\abovecaptionskip}{0.cm}
	\setlength{\belowcaptionskip}{-0.5cm}
	\centering
	\includegraphics[width=1\linewidth]{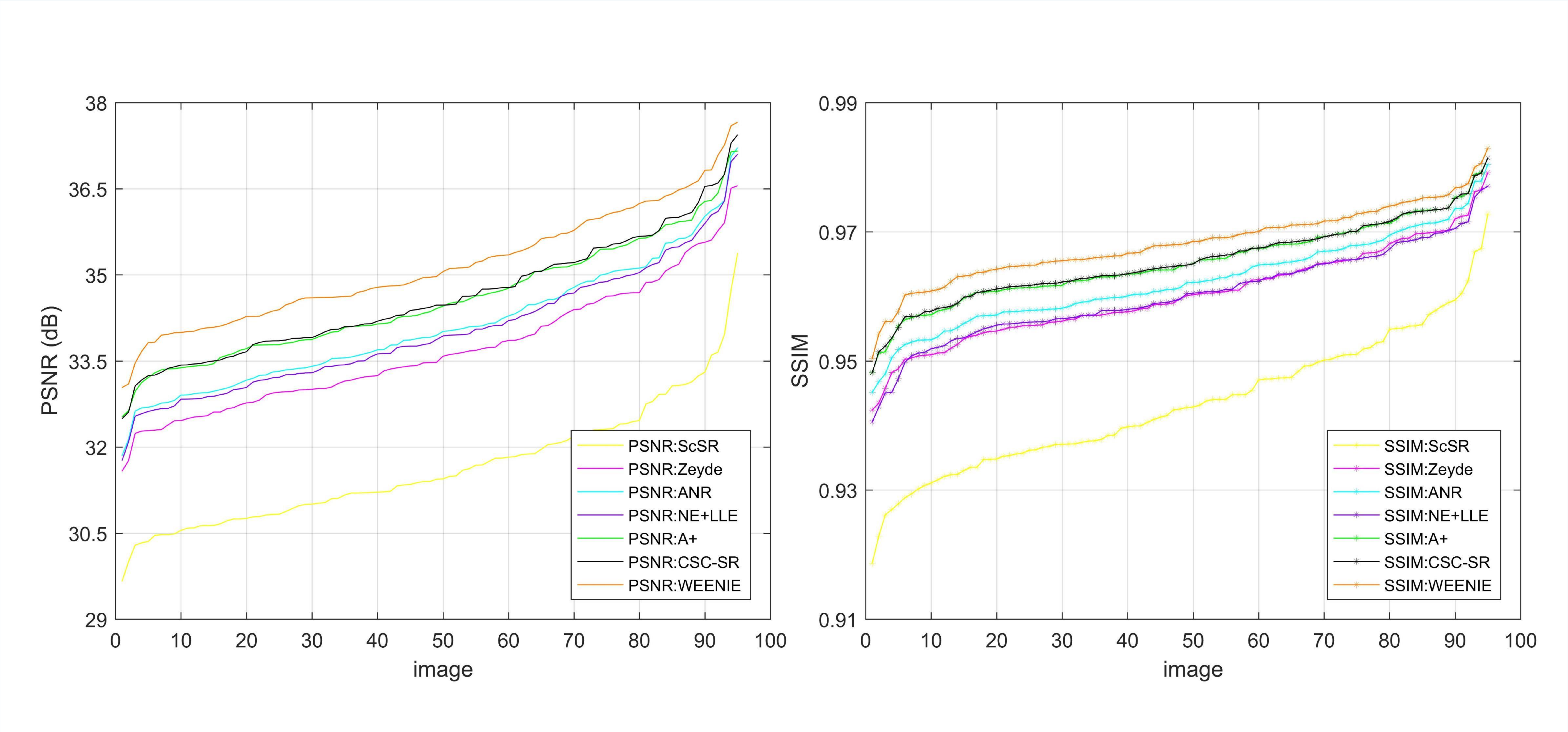}
	\vspace*{-6mm}
	\caption{Performance comparisons of different SR approaches.}
	\label{fig3}
	\vspace*{-2mm}
\end{figure}
\begin{table}
	\begin{center}
		\tiny
		\begin{tabular}{|p{0.68cm}<{\centering}|p{0.54cm}<{\centering}|p{0.58cm}<{\centering}|p{0.515cm}<{\centering}|p{0.705cm}<{\centering}|p{0.368cm}<{\centering}|p{0.9328cm}<{\centering}|p{0.51cm}<{\centering}|}
			\hline
			Metric(avg.) &  ScSR~\cite{yang2010image} & Zeyde~\cite{zeyde2010single} & ANR~\cite{timofte2013anchored} & NE+LLE~\cite{chang2004super} & A+~\cite{timofte2014a+} & CSC-SR~\cite{gu2015convolutional} & WEENIE\\
			\hline\hline
			PSNR(dB) & 31.63 & 33.68 & 34.09 & 34.00 & 34.55 & 34.60 & \textbf{35.13}\\
			\hline
			SSIM & 0.9654 & 0.9623 & 0.9433 & 0.9623 & 0.9591 & 0.9604 & \textbf{0.9681}\\
			\hline
		\end{tabular}
	\end{center}
	\vspace*{-2mm}
	\caption{Quantitative evaluation (PSNR and SSIM): WEENIE vs. other SR methods on 95 subjects of the IXI dataset.}
	\label{tab1}
	\vspace*{-1mm}
\end{table}

For the problem of image SR, we focus on the PD-w subjects of the IXI dataset to compare the proposed WEENIE model with several state-of-the-art SR approaches: sparse coding-based SR method (ScSR)~\cite{yang2010image}, anchored neighborhood regression method (ANR)~\cite{timofte2013anchored}, neighbor embedding + locally linear embedding method (NE+LLE) ~\cite{chang2004super}, Zeyde's method~\cite{zeyde2010single}, convolutional sparse coding-based SR method (CSC-SR)~\cite{gu2015convolutional}, and adjusted anchored neighborhood regression method (A+)~\cite{timofte2014a+}. We perform image SR with scaling factor 2, and show visual results on an example slice in Fig. \ref{fig2}. The quantitative results for different methods are shown in Fig. \ref{fig3}, and the average PSNR and SSIM for all 95 test subjects are listed in Table \ref{tab1}. The proposed method, in the case of brain image SR, obtains the best PSNR and SSIM values. The improvements show that the MMD regularized joint learning property on CSC has more influence than the classic sparse coding-based methods as well as the state-of-the-arts. It states that using MMD combined with the joint CSC indeed improves the representation power of the learned filter pairs.
\begin{figure*}[t]
	\setlength{\abovecaptionskip}{0.cm}
	\setlength{\belowcaptionskip}{-0.5cm}
	\centering
	\includegraphics[width=0.7\linewidth]{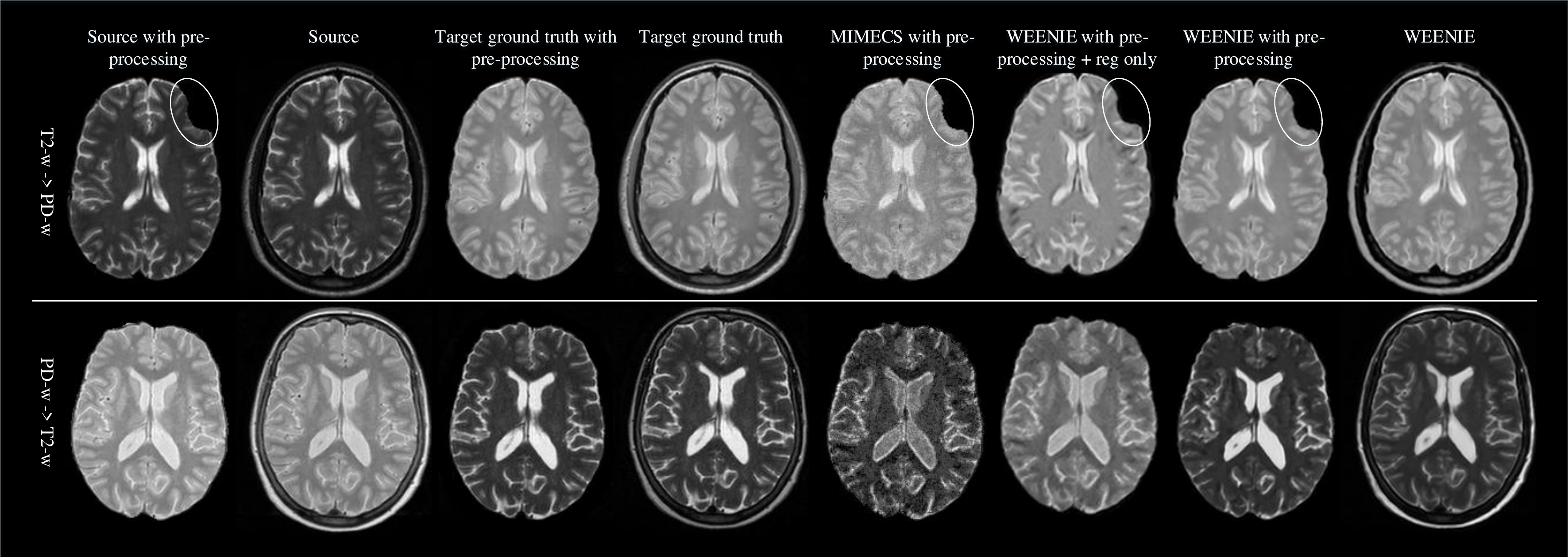}
	\vspace*{-1mm}
	\caption{Visual comparison of synthesized results using different methods.}
	\label{fig4}
	\vspace*{-3mm}
\end{figure*}
\begin{table*}[t]
	\notag
	\begin{center}
		\scriptsize
		\begin{tabular}{|c||c|c||c|c|c||c|c|c|}
			\hline
			\multirow{3}{*}{Metric(avg.)} & \multicolumn{8}{c|}{IXI}\\
			\cline{2-9} & PD$->$T2 & T2$->$PD & \multicolumn{3}{c||}{PD$->$T2+PRE} & \multicolumn{3}{c|}{T2$->$PD+PRE}\\
			\cline{2-9} & \multicolumn{2}{c||}{WEENIE} & MIMECS & WEENIE(reg) & WEENIE & MIMECS & WEENIE(reg) & WEENIE\\
			\hline
			PSNR(dB) & 37.77 & 31.77 & 30.60 & 30.93 & \textbf{33.43} & 29.85 & 30.29 & \textbf{31.00}\\
			\hline
			SSIM & 0.8634 & 0.8575 & 0.7944 & 0.8004 & \textbf{0.8552} & 0.7503 & 0.7612 & \textbf{0.8595}\\
			\hline
		\end{tabular}
	\end{center}
\vspace*{-7mm}
\end{table*}
\begin{table*}[t!]
	\begin{center}
		\scriptsize
	\begin{tabular}{|c||c|c|c|c||c|c|c|c|}
		\hline
		\multirow{3}{*}{Metric(avg.)} & \multicolumn{8}{c|}{NAMIC}\\
		\cline{2-9} & \multicolumn{4}{c||}{T1$->$T2} & \multicolumn{4}{c|}{T2$->$T1}\\
		\cline{2-9} & MIMECS & Ve-US & Ve-S & WEENIE & MIMECS & Ve-US & Ve-S & WEENIE \\
		\hline
		PSNR(dB) & 24.36 & 26.51 & 27.14 & \textbf{27.30} & 27.26 & 27.81 & 29.04 & \textbf{30.35}\\
		\hline
		SSIM & 0.8771 & 0.8874 & 0.8934 & \textbf{0.8983} & 0.9166 & 0.9130 & 0.9173 & \textbf{0.9270}\\
		\hline
	\end{tabular}
	\end{center}
	\vspace*{-3mm}
	\caption{Quantitative evaluation (PSNR and SSIM): WEENIE vs. other synthesis methods on IXI and NAMIC datasets.}
	\label{tab2}
	\vspace*{-5mm}
\end{table*}
\subsection{Simultaneous Super-Resolution and Cross-Modality Synthesis}
\begin{figure}[t]
	\setlength{\abovecaptionskip}{0.cm}
	\setlength{\belowcaptionskip}{-0.5cm}
	\centering
	\includegraphics[width=1\linewidth]{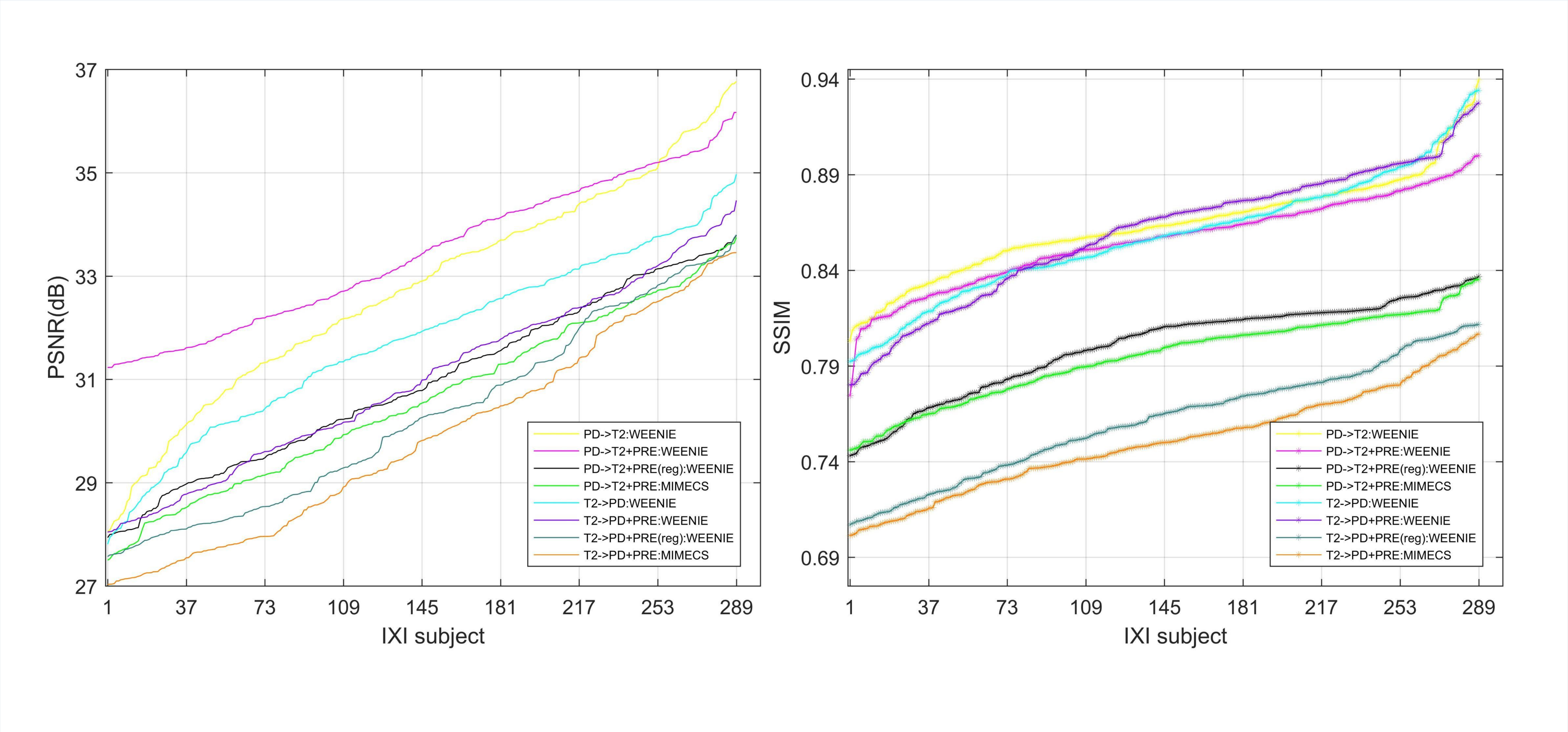}
	\vspace*{-5mm}
	\caption{SRCMS results: WEENIE vs. MIMECS on IXI dataset.}
	\label{fig5}
	\vspace*{-4mm}
\end{figure}
\begin{figure}[t]
	\setlength{\abovecaptionskip}{0.cm}
	\setlength{\belowcaptionskip}{-0.5cm}
	\centering
	\includegraphics[width=1\linewidth]{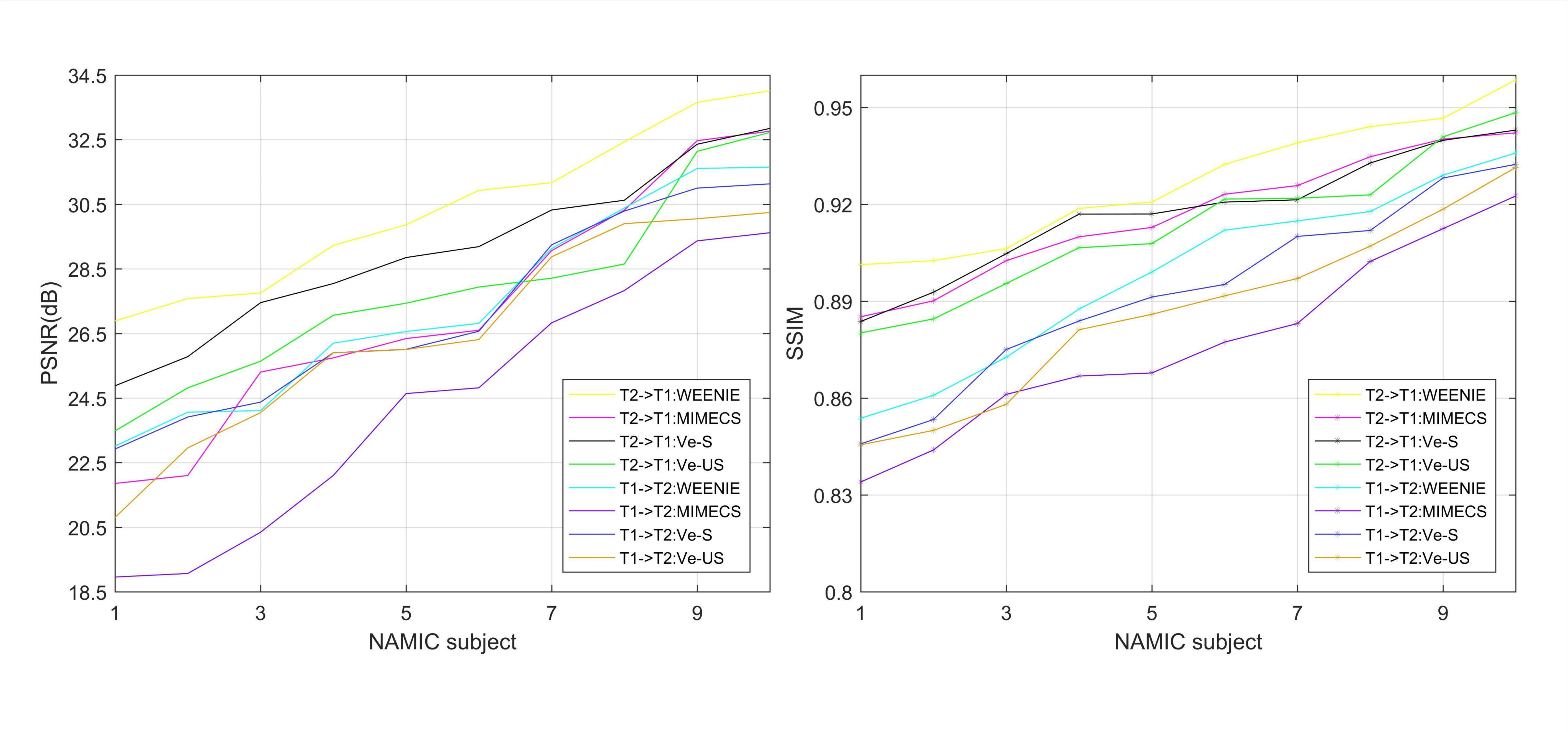}
	\vspace*{-5mm}
	\caption{SRCMS: WEENIE vs. MIMECS on NAMIC dataset.}
	\label{fig6}
	\vspace*{-1mm}
\end{figure}
To comprehensively test the robustness of the proposed WEENIE method, we perform SRCMS on both datasets involving six groups of experiments: (1) synthesizing SR T2-w image from LR PD-w acquisition and (2) \textit{vice versa}; (3) generating SR T2-w image from LR PD-w input based on pre-processed data (\ie, skull strapping and bias corrections) and (4) \textit{vice versa}; (5) synthesizing SR T1-w image from LR T2-w subject and (6) \textit{vice versa}. The first four sets of experiments are conducted on the IXI dataset while the last two cases are evaluated on the NAMIC dataset. The state-of-the-art synthesis methods include Vemulapalli's supervised approach (V-S)~\cite{vemulapalli2015unsupervised}, Vemulapalli's unsupervised approach (V-US)~\cite{vemulapalli2015unsupervised} and MR image exampled-based contrast synthesis (MIMECS)~\cite{roy2013magnetic} approach. However, Vemulapalli's methods cannot be applied for our problem, because they only contain the cross-modality synthesis stage used in the NAMIC dataset. Original data (without degradation processing) are used in all Vemulapalli’s methods. MIMECS takes image SR into mind and adopts two independent steps (\ie synthesis+SR) to solve the problem. We compare our results on
only using registered image pairs denoted by WEENIE(reg) (that can directly substantiate the benefits of involving unpaired data) and the results using all training images with/without preprocessing for the proposed method against MIMECS, V-US and V-S in above six cases and demonstrate examples in Fig. \ref{fig4} for visual inspection. The advantage of our method over the MIMECS shows, \eg, in white matter structures, as well as in the overall intensity profile. We show the quantitative results in Fig. \ref{fig5}, and Fig. \ref{fig6}, and summarize the averaged values in Table \ref{tab2}, respectively. It can be seen that the performance of our algorithm is consistent across two whole datasets, reaching the best PSNR and SSIM for almost all subjects.
\vspace*{-1mm}
\section{Conclusion}
\vspace*{-1mm}
In this paper, we proposed a novel weakly-supervised joint convolutional sparse coding (WEENIE) method for simultaneous super-resolution and cross-modality synthesis (SRCMS) in 3D MRI. Different from conventional joint learning approaches based on sparse representation in supervised setting, WEENIE only requires a small set of registered image pairs and automatically aligns the correspondence for auxiliary unpaired images to span the diversities of the original learning system. By means of the designed hetero-domain alignment term, a set of filter pairs and the mapping function were jointly optimized in a common feature space. Furthermore, we integrated our model with a divergence minimization term to enhance robustness. With the benefit of consistency prior, WEENIE directly employs the whole image, which naturally captures the correlation between local neighborhoods. As a result, the proposed method can be applied to both brain image SR and SRCMS problems. Extensive results showed that WEENIE can achieve superior performance against state-of-the-art methods.

{\small
\bibliographystyle{ieee}
\bibliography{egbib}

\begin{thebibliography}{10}\itemsep=-1pt

\bibitem{bahrami2015hierarchical}
K.~Bahrami, F.~Shi, X.~Zong, H.~W. Shin, H.~An, and D.~Shen.
\newblock Hierarchical reconstruction of 7t-like images from 3t mri using
  multi-level cca and group sparsity.
\newblock In {\em International Conference on Medical Image Computing and
  Computer-Assisted Intervention}, pages 659--666. Springer, 2015.

\bibitem{bristow2013fast}
H.~Bristow, A.~Eriksson, and S.~Lucey.
\newblock Fast convolutional sparse coding.
\newblock In {\em Proceedings of the IEEE Conference on Computer Vision and
  Pattern Recognition}, pages 391--398, 2013.

\bibitem{chang2004super}
H.~Chang, D.-Y. Yeung, and Y.~Xiong.
\newblock Super-resolution through neighbor embedding.
\newblock In {\em Computer Vision and Pattern Recognition, 2004. CVPR 2004.
  Proceedings of the 2004 IEEE Computer Society Conference on}, volume~1, pages
  I--I. IEEE, 2004.

\bibitem{chen2014recognizing}
L.~Chen, W.~Li, and D.~Xu.
\newblock Recognizing rgb images by learning from rgb-d data.
\newblock In {\em Proceedings of the IEEE Conference on Computer Vision and
  Pattern Recognition}, pages 1418--1425, 2014.

\bibitem{dong2016image}
C.~Dong, C.~C. Loy, K.~He, and X.~Tang.
\newblock Image super-resolution using deep convolutional networks.
\newblock {\em IEEE transactions on pattern analysis and machine intelligence},
  38(2):295--307, 2016.

\bibitem{efros2001image}
A.~A. Efros and W.~T. Freeman.
\newblock Image quilting for texture synthesis and transfer.
\newblock In {\em Proceedings of the 28th annual conference on Computer
  graphics and interactive techniques}, pages 341--346. ACM, 2001.

\bibitem{freeman2002example}
W.~T. Freeman, T.~R. Jones, and E.~C. Pasztor.
\newblock Example-based super-resolution.
\newblock {\em IEEE Computer graphics and Applications}, 22(2):56--65, 2002.

\bibitem{freeman2000learning}
W.~T. Freeman, E.~C. Pasztor, and O.~T. Carmichael.
\newblock Learning low-level vision.
\newblock {\em International journal of computer vision}, 40(1):25--47, 2000.

\bibitem{gao2012face}
X.~Gao, N.~Wang, D.~Tao, and X.~Li.
\newblock Face sketch--photo synthesis and retrieval using sparse
  representation.
\newblock {\em IEEE Transactions on circuits and systems for video technology},
  22(8):1213--1226, 2012.

\bibitem{gatys2016image}
L.~A. Gatys, A.~S. Ecker, and M.~Bethge.
\newblock Image style transfer using convolutional neural networks.
\newblock In {\em Proceedings of the IEEE Conference on Computer Vision and
  Pattern Recognition}, pages 2414--2423, 2016.

\bibitem{gu2015convolutional}
S.~Gu, W.~Zuo, Q.~Xie, D.~Meng, X.~Feng, and L.~Zhang.
\newblock Convolutional sparse coding for image super-resolution.
\newblock In {\em Proceedings of the IEEE International Conference on Computer
  Vision}, pages 1823--1831, 2015.

\bibitem{heide2015fast}
F.~Heide, W.~Heidrich, and G.~Wetzstein.
\newblock Fast and flexible convolutional sparse coding.
\newblock In {\em 2015 IEEE Conference on Computer Vision and Pattern
  Recognition (CVPR)}, pages 5135--5143. IEEE, 2015.

\bibitem{hertzmann2001image}
A.~Hertzmann, C.~E. Jacobs, N.~Oliver, B.~Curless, and D.~H. Salesin.
\newblock Image analogies.
\newblock In {\em Proceedings of the 28th annual conference on Computer
  graphics and interactive techniques}, pages 327--340. ACM, 2001.

\bibitem{huang2013coupled}
D.-A. Huang and Y.-C. Frank~Wang.
\newblock Coupled dictionary and feature space learning with applications to
  cross-domain image synthesis and recognition.
\newblock In {\em Proceedings of the IEEE international conference on computer
  vision}, pages 2496--2503, 2013.

\bibitem{huang2015single}
J.-B. Huang, A.~Singh, and N.~Ahuja.
\newblock Single image super-resolution from transformed self-exemplars.
\newblock In {\em 2015 IEEE Conference on Computer Vision and Pattern
  Recognition (CVPR)}, pages 5197--5206. IEEE, 2015.

\bibitem{huang2016color}
Y.~Huang, F.~Zhu, L.~Shao, and A.~F. Frangi.
\newblock Color object recognition via cross-domain learning on rgb-d images.
\newblock In {\em Robotics and Automation (ICRA), 2016 IEEE International
  Conference on}, pages 1672--1677. IEEE, 2016.

\bibitem{iglesias2013synthesizing}
J.~E. Iglesias, E.~Konukoglu, D.~Zikic, B.~Glocker, K.~Van~Leemput, and
  B.~Fischl.
\newblock Is synthesizing mri contrast useful for inter-modality analysis?
\newblock In {\em International Conference on Medical Image Computing and
  Computer-Assisted Intervention}, pages 631--638. Springer, 2013.

\bibitem{jia2013image}
K.~Jia, X.~Wang, and X.~Tang.
\newblock Image transformation based on learning dictionaries across image
  spaces.
\newblock {\em IEEE transactions on pattern analysis and machine intelligence},
  35(2):367--380, 2013.

\bibitem{jog2013magnetic}
A.~Jog, S.~Roy, A.~Carass, and J.~L. Prince.
\newblock Magnetic resonance image synthesis through patch regression.
\newblock In {\em 2013 IEEE 10th International Symposium on Biomedical
  Imaging}, pages 350--353. IEEE, 2013.

\bibitem{keys1981cubic}
R.~Keys.
\newblock Cubic convolution interpolation for digital image processing.
\newblock {\em IEEE transactions on acoustics, speech, and signal processing},
  29(6):1153--1160, 1981.

\bibitem{li2001new}
X.~Li and M.~T. Orchard.
\newblock New edge-directed interpolation.
\newblock {\em IEEE transactions on image processing}, 10(10):1521--1527, 2001.

\bibitem{long2013transfer}
M.~Long, G.~Ding, J.~Wang, J.~Sun, Y.~Guo, and P.~S. Yu.
\newblock Transfer sparse coding for robust image representation.
\newblock In {\em Proceedings of the IEEE conference on computer vision and
  pattern recognition}, pages 407--414, 2013.

\bibitem{monay2007modeling}
F.~Monay and D.~Gatica-Perez.
\newblock Modeling semantic aspects for cross-media image indexing.
\newblock {\em IEEE Transactions on Pattern Analysis and Machine Intelligence},
  29(10):1802--1817, 2007.

\bibitem{mueller2005alzheimer}
S.~G. Mueller, M.~W. Weiner, L.~J. Thal, R.~C. Petersen, C.~Jack, W.~Jagust,
  J.~Q. Trojanowski, A.~W. Toga, and L.~Beckett.
\newblock The alzheimer's disease neuroimaging initiative.
\newblock {\em Neuroimaging Clinics of North America}, 15(4):869--877, 2005.

\bibitem{rousseau2010non}
F.~Rousseau, A.~D.~N. Initiative, et~al.
\newblock A non-local approach for image super-resolution using intermodality
  priors.
\newblock {\em Medical image analysis}, 14(4):594--605, 2010.

\bibitem{roy2013magnetic}
S.~Roy, A.~Carass, and J.~L. Prince.
\newblock Magnetic resonance image example-based contrast synthesis.
\newblock {\em IEEE transactions on medical imaging}, 32(12):2348--2363, 2013.

\bibitem{rueda2013single}
A.~Rueda, N.~Malpica, and E.~Romero.
\newblock Single-image super-resolution of brain mr images using overcomplete
  dictionaries.
\newblock {\em Medical image analysis}, 17(1):113--132, 2013.

\bibitem{shao2007order}
L.~Shao and M.~Zhao.
\newblock Order statistic filters for image interpolation.
\newblock In {\em 2007 IEEE International Conference on Multimedia and Expo},
  pages 452--455. IEEE, 2007.

\bibitem{smeulders2000content}
A.~W. Smeulders, M.~Worring, S.~Santini, A.~Gupta, and R.~Jain.
\newblock Content-based image retrieval at the end of the early years.
\newblock {\em IEEE Transactions on pattern analysis and machine intelligence},
  22(12):1349--1380, 2000.

\bibitem{tang2017pairwise}
Y.~Tang and L.~Shao.
\newblock Pairwise operator learning for patch-based single-image
  super-resolution.
\newblock {\em IEEE Transactions on Image Processing}, 26(2):994--1003, 2017.

\bibitem{timofte2013anchored}
R.~Timofte, V.~De~Smet, and L.~Van~Gool.
\newblock Anchored neighborhood regression for fast example-based
  super-resolution.
\newblock In {\em Proceedings of the IEEE International Conference on Computer
  Vision}, pages 1920--1927, 2013.

\bibitem{timofte2014a+}
R.~Timofte, V.~De~Smet, and L.~Van~Gool.
\newblock A+: Adjusted anchored neighborhood regression for fast
  super-resolution.
\newblock In {\em Asian Conference on Computer Vision}, pages 111--126.
  Springer, 2014.

\bibitem{van2015cross}
H.~Van~Nguyen, K.~Zhou, and R.~Vemulapalli.
\newblock Cross-domain synthesis of medical images using efficient
  location-sensitive deep network.
\newblock In {\em International Conference on Medical Image Computing and
  Computer-Assisted Intervention}, pages 677--684. Springer, 2015.

\bibitem{vemulapalli2015unsupervised}
R.~Vemulapalli, H.~Van~Nguyen, and S.~Kevin~Zhou.
\newblock Unsupervised cross-modal synthesis of subject-specific scans.
\newblock In {\em Proceedings of the IEEE International Conference on Computer
  Vision}, pages 630--638, 2015.

\bibitem{wang2012semi}
S.~Wang, L.~Zhang, Y.~Liang, and Q.~Pan.
\newblock Semi-coupled dictionary learning with applications to image
  super-resolution and photo-sketch synthesis.
\newblock In {\em Computer Vision and Pattern Recognition (CVPR), 2012 IEEE
  Conference on}, pages 2216--2223. IEEE, 2012.

\bibitem{wang2009face}
X.~Wang and X.~Tang.
\newblock Face photo-sketch synthesis and recognition.
\newblock {\em IEEE Transactions on Pattern Analysis and Machine Intelligence},
  31(11):1955--1967, 2009.

\bibitem{wang2004image}
Z.~Wang, A.~C. Bovik, H.~R. Sheikh, and E.~P. Simoncelli.
\newblock Image quality assessment: from error visibility to structural
  similarity.
\newblock {\em IEEE transactions on image processing}, 13(4):600--612, 2004.

\bibitem{yang2010image}
J.~Yang, J.~Wright, T.~S. Huang, and Y.~Ma.
\newblock Image super-resolution via sparse representation.
\newblock {\em IEEE transactions on image processing}, 19(11):2861--2873, 2010.

\bibitem{zeiler2010deconvolutional}
M.~D. Zeiler, D.~Krishnan, G.~W. Taylor, and R.~Fergus.
\newblock Deconvolutional networks.
\newblock In {\em Computer Vision and Pattern Recognition (CVPR), 2010 IEEE
  Conference on}, pages 2528--2535. IEEE, 2010.

\bibitem{zeyde2010single}
R.~Zeyde, M.~Elad, and M.~Protter.
\newblock On single image scale-up using sparse-representations.
\newblock In {\em International conference on curves and surfaces}, pages
  711--730. Springer, 2010.

\bibitem{zhang2006edge}
L.~Zhang and X.~Wu.
\newblock An edge-guided image interpolation algorithm via directional
  filtering and data fusion.
\newblock {\em IEEE transactions on Image Processing}, 15(8):2226--2238, 2006.

\bibitem{zheng2016hetero}
F.~Zheng, Y.~Tang, and L.~Shao.
\newblock Hetero-manifold regularisation for cross-modal hashing.
\newblock {\em IEEE Transactions on Pattern Analysis and Machine Intelligence},
  2016.

\end{thebibliography}
}

\end{document}